\documentclass{article}
\usepackage{spconf,amsmath,graphicx}
\usepackage{url}
\usepackage{amssymb}
\usepackage{epsfig}
\usepackage{balance}
\usepackage{booktabs}
\usepackage{cite}
\usepackage{multirow}
\usepackage{tikz}
\usepackage{marvosym}
\usepackage{ifsym}

\title{Explicit and Implicit Knowledge Distillation via unlabeled data}

\name{Yuzheng Wang, Zuhao Ge, Zhaoyu Chen, Xian Liu, Chuangjia Ma, Yunquan Sun$^{\textrm{\Letter}}$, Lizhe Qi$^{\textrm{\Letter}}$\thanks{\textrm{\Letter}\;\;The corresponding authors are Lizhe Qi and Yunquan Sun. This work is supported by Natural Science Foundation of Jiangxi Province (No.20212BAB202026), Shanghai Municipal Science and Technology Major Project (No.2021SHZDZX0103), the Shanghai Engineering Research Center of AI \& Robotics, Fudan University, China, and the Engineering Research Center of AI \& Robotics, Ministry of Education, China.}}

\address{Academy for Engineering \& Technology, Fudan University, Shanghai, China}
\begin{document}
\maketitle

\begin{abstract}
Data-free knowledge distillation is a challenging model lightweight task for scenarios in which the original dataset is not available.
Previous methods require a lot of extra computational costs to update one or more generators and their naive imitate-learning lead to lower distillation efficiency.
Based on these observations, we first propose an efficient unlabeled sample selection method to replace high computational generators and focus on improving the training efficiency of the selected samples.
Then, a class-dropping mechanism is designed to suppress the label noise caused by the data domain shifts.
Finally, we propose a distillation method that incorporates explicit features and implicit structured relations to improve the effect of distillation.
Experimental results show that our method can quickly converge and obtain higher accuracy than other state-of-the-art methods.
\end{abstract}


\section{Introduction}
\label{sec:intro}

Deep neural networks are gradually developing toward large-scale models \cite{devlin2018bert,karras2019style,yang2023target,liu2022learninga,liu2022appearance,Chen_2022_CVPR}. The changes have brought about an impressive technological breakthrough \cite{goodfellow2014generative,chen2022shape,liu2022efficient,9428443,yang2022disentangled,yang2022learning,wang2022boosting,9987686,huang2022cmua,wang2023adversarial}, but applying these technologies to mobile devices such as mobile phones, driverless cars, and tiny robots is difficult. 
Besides, the source data cannot be obtained in many cases due to data security, such as fingerprints, faces, and medical records images.
Therefore, model compression and data-free technology are the keys to breaking barriers.
In this situation, Data-Free Knowledge Distillation (DFKD) is proposed\cite{lopes2017data}.
In this process, an easy-to-deploy lightweight student model is trained with the help of redundant teacher models without original training data, which is much more efficient than retraining models. 
Therefore, it is widely used in various fields and has developed rapidly in recent years.

\begin{figure}[t]
	\centering
	\includegraphics[scale=0.278]{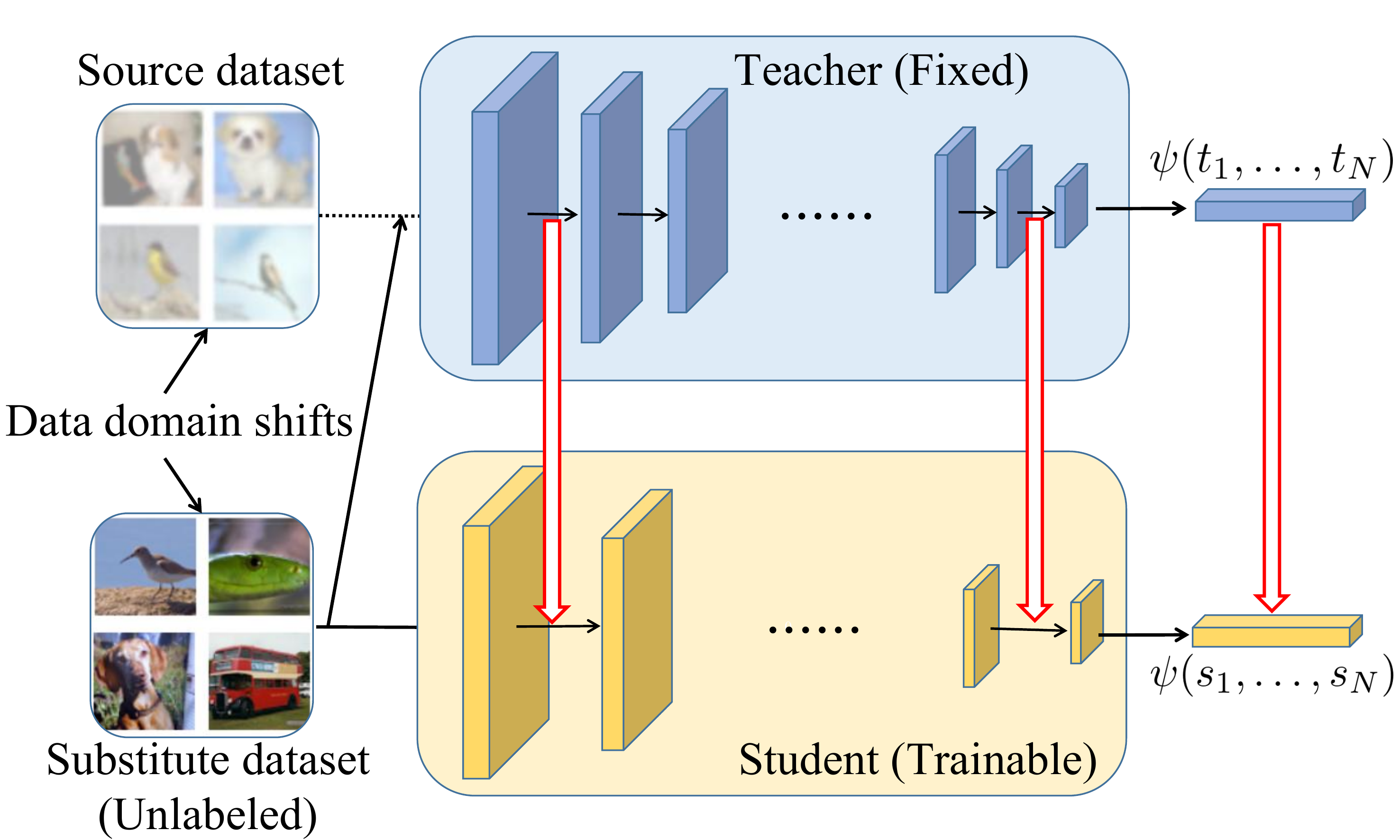}
	\vspace{-0.1cm}
	\caption{The pipeline of our method. Only the teacher model and the unlabeled substitute dataset are available during training. The red arrows denote our proposed explicit and implicit distillation losses.}
	\label{fig1}
	\vspace{-0.3cm}
\end{figure}

There are currently two ideas in the DFKD field. 
One idea is to set up a generation module to supplement the training data. 
Chen \emph{et al}.\cite{chen2019} combine knowledge distillation with Generative Adversarial Networks (GANs). Fang \emph{et al}.\cite{fang2019} introduce a model difference to force the generator to produce more complex samples.
Micaelli \emph{et al}.\cite{micaelli2019zero} use the generated samples that can confuse the discriminator to make student learning more efficient.
Fang \emph{et al}.\cite{fang2021up} propose a local sharing method to reduce the cost of data generation.
Besides, Yin \emph{et al}.\cite{yin2020dreaming} and choi \emph{et al}.\cite{choi2020data} propose a method based on model inversion of teacher network to synthesize more realistic samples. 
Fang \emph{et al}.\cite{fang2021contrastive} propose a method of combining distillation with other compression technologies and achieving extensive results.
Another idea is to use unlabeled substitute data.
Chen \emph{et al}.\cite{chen2021learning} propose selecting samples in the wild without generation module. 
The wild dataset represents a substitute dataset that is easily accessible while ignoring labels, such as the ImageNet dataset \cite{deng2009imagenet}.

Despite encouraging performance, firstly, the methods based on the generation module will generate a large amount of additional computational costs and parameters.
The method based on unlabeled sample selection can avoid these problems. 
However, the previous selection mechanism ignores the amount of information on unlabeled samples, representing the effectiveness of students learning.
Secondly, the training dataset is composed of unlabeled data or random noise transform and lacks supervision information, so it contains a large amount of label noise.
However, the previous methods ignore the disturbance of the noise.
Finally, previous methods force the student to mimic the outputs of a particular data example represented by the teacher, resulting in low convergence speed and lack of a structured knowledge representation, which affects students' performance.

To tackle these issues, we consider a low computational and low noise efficient distillation framework called Efficient Explicit and Implicit Knowledge Distillation (EEIKD).
Specifically, we design an adaptive threshold selection module to avoid additional generation costs.
To suppress the sample noise, we design a class-dropping mechanism, which hardly adds additional computation.
To increase the convergence speed and explore the relationship between multiple samples, we propose a distillation method combining explicit and implicit knowledge, as shown in Fig.\;\ref{fig1}.
The primary contributions and experiments are
summarized below:
\vspace{-0.1cm}
\begin{itemize}
\item We propose an Efficient Explicit and Implicit Knowledge Distillation method, which selects unlabeled substitute samples without additional generation module calculation and parameter costs.
\vspace{-0.1cm}
\item To find more efficient samples from the substitute dataset, we propose an adaptive threshold selection module, which comprehensively considers unlabeled samples' confidence and information content.
\vspace{-0.1cm}
\item We design a lightweight class-dropping mechanism to suppress label noise. Then, we combine explicit and implicit knowledge, which significantly improves the convergence speed and learning efficiency.
\vspace{-0.1cm}
\item Experimental results show that our EEIKD method significantly improves students' performance compared with previous state-of-the-art DFKD methods.
\end{itemize}

\section{METHODOLOGY}

In this section, we first introduce an adaptive threshold module for unlabeled data selection. Then we introduce a label noise suppression mechanism to deal with data domain shifts. Finally, an efficient knowledge distillation method is proposed to train an impressive student.

\subsection{Adaptive Threshold Selection}

To discard unnecessary generator costs, we propose an unlabeled data selection method.
At the same time, we try to select suitable samples to enhance learning efficiency.
On the one hand, the high confidence prediction of the teacher network for an unlabeled sample means that it comprehends the sample better, which helps improve the utilization of samples.
On the other hand, higher confidence means that the prediction of the teacher network is closer to one-hot encoding. 
Compared with the soft target, its prediction has lower entropy and can provide less training information than the soft target \cite{hinton2015distilling}, thus reducing the sample utilization efficiency.
Here, we design a mechanism to balance the two parts.

We denote the unlabeled input sample as $x$, the number of classes as $n$, the unlabeled substitute dataset as $\mathcal{X}$ ($x \in \mathcal{X}$), the candidate dataset as ${\mathcal{X}}'$ and final student training dataset after selection as ${\mathcal{X}}''$.

\vspace{-0.3cm}
$$
\delta=\frac{1}{n^{\gamma}},\;  \underset{x}{\arg\max} \; S(f_{T}(x)) >\delta,\;  \; x \rightarrow \mathcal{X}^{\prime},   \eqno{(1)}
$$

\noindent where $\delta$ is the adaptive threshold, $\gamma$ is a hyperparameter satisfied 0\textless $\gamma$\textless 1, $S$ is the softmax function, and $f_{T}(x)$ is the prediction of teacher network.
Then $ns$ samples are selected in ${\mathcal{X}}'$ to form training set ${\mathcal{X}}''$. 
As $\gamma$ increases and $\delta$ decreases, samples with more information are valued. 
As $\gamma$ decreases and $\delta$ increases, more confident samples are selected.
Through the adaptive threshold setting, samples with low confidence or too little information will not be selected.
Therefore, efficient samples are finally selected, which can better help the student perform well.

\begin{figure*}[t]
	\centering
	\includegraphics[scale=0.53]{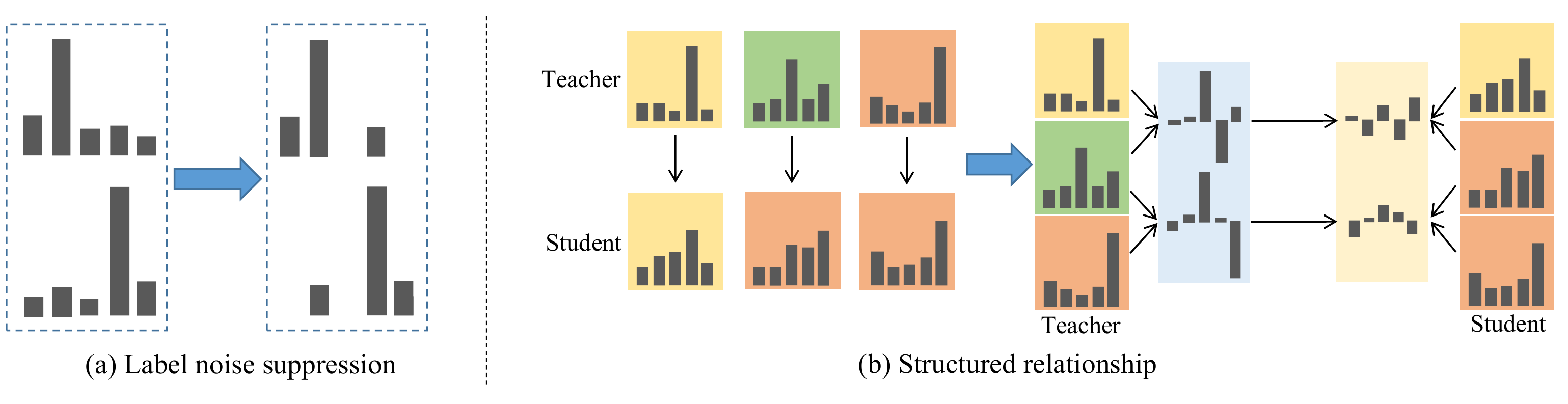}
	\vspace{-0.4cm}
	\caption{The columns denote the confidence prediction. The blue arrows show the difference between the past and our method. (a) Label noise suppression based on the class-dropping module. (b) An implicit structured relation distillation method.}
	\label{fig2}
	\vspace{-0.3cm}
\end{figure*}

\vspace{-0.1cm}
\subsection{Class-Dropping Noise Suppression}
Since classes differ between the unavailable original dataset and the unlabeled substitute dataset, we propose a class-dropping noise suppression module to face the datasets' domain shifts and improve the learning effect of the student network.
The predictions given by the teacher network for the classes with low confidence are often influenced by the shifts between the datasets' domains. 
First, the prediction of these parts will not positively affect the final results.
Further, it also affects the learning efficiency of complex samples as noise. 
Here, we propose a simple yet effective method of confidence mask to suppress the noise from the unlabeled data domain shifts shown in Fig.\;\ref{fig2}(a).

We denote the class-dropping rate as $\alpha$ (0\textless $\alpha$\textless 1), a class as $c$, the mask for class $c$ in a sample as $m_{c}$, the confidence for class $c$ in a sample as $p_{c}$ and the prediction confidence for a sample as $P={\{p_{1},p_{2},\dots, p_{n}\}}$. 
$K$ is the number of classes reserved, and $K$ is equal to $\lfloor (1-\alpha) \times n \rfloor$.
$$m_{c}=\begin{cases}
 1, & \text{ if    }  p_{c} \geq \text{top-} K (P), \\
 0, & \text{ otherwise, } 
\end{cases} \eqno{(2)}$$
\vspace{-0.3cm}
$$M_{x}=\left \{  m_{1},  m_{2},\cdots , m_{n} \right \},\; \hat{f_{T}}(x) =f_{T}(x)\odot M_{x}, \eqno{(3)}$$
\vspace{-0.3cm}

\noindent where $M_{x}$ is the sample confidence mask and $\text{top-}K(P)$ is the prediction confidence of the $K$-th largest class of a sample. Then the masked confidence matrix $\hat{f_{T}}(x)$ can be obtained.
When calculating the constraints of the final output between the teacher and student, the masked confidence matrix is used to replace the complete output.
The student will learn a low-noise representation to suppress noise.

\vspace{-0.1cm}
\subsection{Explicit and Implicit Knowledge Distillation}

\subsubsection{Knowledge distillation loss}
Knowledge distillation \cite{hinton2015distilling} is an important model compression technique. $f_{T}(x)$ and $f_{S}(x)$ denote the output of teacher network and student network. Knowledge distillation loss is expressed to minimize objective function:
$$\mathcal{L}_{K\!D}=\sum_{x \in \mathcal{X}} D_{K\!L}\left(S\left(\frac{f_{T}(x)}{\tau}\right), S\left(\frac{f_{S}(x)}{\tau}\right)\right), \eqno{(4)}$$where $D_{K\!L}$ is the Kullback-Leibler divergence and $\tau$ is the distillation temperature. 
The knowledge distillation loss $\mathcal{L}_{K\!D}$ allows the student to imitate the teacher's output.
However, $\mathcal{L}_{K\!D}$ is usually not particularly efficient \cite{yim2017gift}, especially facing the datasets domain shifts.

\subsubsection{Explicit feature distillation}
In multi-layer neural networks, the output of the lower layer is locally concerned with texture information. The high-level output vision is gradually increasing, and the global information is increasingly focused. 
Here we choose the output after the first Batch Normalization (BN) layer \cite{ioffe2015batch} and the input before the final linear layer. The former focuses on local texture features and preserves the features of training pictures which is helpful for the rapid convergence of the student network. 
The latter is directly related to the final effect. The attention distillation loss is described as:
\vspace{-0.1cm}
$$\mathcal{L}_{AT\!\_f}=E_{x \sim P_{\text {data }}(x)}\left\|f_{T\!\_{f}}(x)-f_{S\!\_{f}}(x)\right\|_{1}, \eqno{(5)}$$
\vspace{-0.2cm}
$$\mathcal{L}_{AT\!\_{b}}=E_{x \sim P_{\text {data }}(x)}\left\|f_{T\!\_{b}}(x)-f_{S\!\_{b}}(x)\right\|_{1}, \eqno{(6)}$$
\vspace{-0.2cm}
$$\mathcal{L}_{AT}=\mathcal{L}_{AT\!\_{f}}+\mathcal{L}_{A T\!\_{b}}, \eqno{(7)}$$
\vspace{-0.3cm}

\noindent where $\left \| \cdot  \right \| _{1}$ is the $\ell 1$ norm, $f_{T\!\_f}(x)$ and $f_{T\!\_b}(x)$ is the output after the first BN layer and the input before the final linear layer of the teacher, $f_{S\!\_f}(x)$ and $f_{S\!\_b}(x)$ is the output of the two layers of the student.
We believe that we can obtain a faster convergence speed by learning these features. The relevant experimental verification is in the next section.
%

\subsubsection{Implicit structured relation distillation}

In the process of model learning, the learning effect of different classes is usually different.
It is challenging to learn complex classes directly but relatively easy to learn the differences in sample confidence distribution in a mini-batch.
We aim to make the learning of complex classes efficiently through a structured relational distillation shown in Fig.\;\ref{fig2}(b). 
The structured relation denotes the connection between multiple samples rather than a single sample example.


We denote the batch size as \emph{N}, the structured differentiation relationships as $\psi $, teacher's predictions as $\hat{f_{T}}(N) \!=\! \left(t_{1}, \ldots, t_{N}\right)$ and the structured differentiation loss as $\mathcal{L}_{D}$.
The implicit structured distillation calculation is as follows:
\vspace{-0.15cm}
$$\psi\left(t_{i}\right) \!= \! \frac{1}{N-1}\! \sum_{j=1, j \neq i}^{N} \! \left\|t_{i}-t_{j}\right\|_{2}, \; \xi_{t}\! =\! \frac{1}{N} \! \sum_{i=1}^{N} \! \psi\left(t_{i}\right){, } \eqno{(8)}$$
\vspace{-0.15cm}
$$\mathcal{L}_{D}=\sum_{i=1}^{N} \ell_{\delta}\left(\frac{\psi\left(t_{i}\right)}{\xi_{t}}, \frac{\psi\left(s_{i}\right)}{\xi_{s}}\right){,} \eqno{(9)}$$where $\ell_{\delta}$ is the Huber loss. The structured differentiation relationships of the student $\psi(s_i)$ are similar to Eq.\;8.
Finally, we can get the total loss by summing up all losses as:
\vspace{-0.0cm}
$$\mathcal{L}_{total}=\mathcal{L}_{K\!D}+\lambda_{1}\! \cdot \!\mathcal{L}_{AT}+\lambda_{2}\! \cdot \!\mathcal{L}_{D}, \eqno{(10)}$$
\vspace{-0.3cm}

\noindent where $\lambda_{1}$, $\lambda_{2}$ are the loss trade-off parameters.

\section{EXPERIMENTS}
In this section, we first verify the effectiveness of our proposed method through the hyperparametric and ablation experiments. Then we compare it with current state-of-the-art methods to prove its superiority.

\subsection{Experimental Settings}

\noindent\textbf{Datasets:} Our setting is selecting the samples that can better help the student learn from the unlabeled substitute dataset to replace the unavailable source dataset following DFND \cite{chen2021learning}. \noindent\textbf{Unavailable source dataset:} 32$\times $32 CIFAR-10 and CIFAR-100 \cite{krizhevsky2009learning} contain 50K training and 10K testing datasets from 10 and 100 classes. \noindent \textbf{Unlabeled substitute dataset:} ImageNet dataset \cite{deng2009imagenet}. The ImageNet dataset is resized to 32$\times $32 to meet the input requirements of the original model.

\noindent\textbf{Implementation Details:} The proposed method is implemented in PyTorch \cite{paszke2019pytorch} and trained with eight RTX 2080 Ti GPUs. In the comparative experiment, we expect two groups of experiments to meet the needs of different situations (Tiny or Large). We select ResNet-34 \cite{he2016deep} as the teacher network and ResNet-18 \cite{he2016deep} as the student network following the past baseline. Then we choose 150K and 500K samples for Tiny and Large schemes and train for 200 and 800 epochs. For the DFND, we keep the 600K samples and 800 epochs from the original paper, and the student model is trained more times than our method. 
Finally, we choose $\lambda_{1}$ as 0.1 and $\lambda_{2}$ as 1, use the SGD optimizer with the momentum as 0.9, weight decay as $5\times10^{-4}$, and the learning rate initially equal to 0.1.

\begin{table}[h]
\vspace{-0.3cm}
\centering
\caption{Student accuracy (\%) about parameter experiments on adaptive threshold $\gamma$ and class-dropping rate $\alpha$.}
\label{tab1}
\vspace{0.1cm}
\setlength{\tabcolsep}{0.95mm}
\scalebox{0.85}{
\begin{tabular}{@{}cccc|cccc@{}}
\toprule
ID & $\gamma$ & CIFAR-10       & CIFAR-100      & ID  & $\alpha$ & CIFAR-10       & CIFAR-100      \\ \midrule
1 & 0.05     & 94.24          & 76.36          & 6  & 0        & 94.49          & 76.44          \\
2 & 0.1      & \textbf{94.49} & \textbf{76.44} & 7  & 0.3      & \textbf{94.57} & 76.47          \\
3 & 0.2      & 94.13          & 75.77          & 8  & 0.5      & 94.35          & \textbf{76.96} \\
4 & 0.3      & 92.56          & 75.25          & 9  & 0.7      & 92.31          & 75.25          \\
5 & 0.5      & 92.36          & 74.41          & 10 & 0.9      & 72.91          & 67.25          \\ \bottomrule
\end{tabular}
}
\vspace{-0.3cm}
\end{table}

\begin{table}[h]
\vspace{-0.3cm}
\centering
\caption{Ablation experiments on CIFAR-100 with selected 300K samples at different epochs.}
\label{tab2}
\vspace{0.1cm}
\setlength{\tabcolsep}{3mm}
\scalebox{0.8}{
\begin{tabular}{c|cccc}
\hline
\multirow{2}{*}{Method}               & \multicolumn{4}{c}{Epochs}    \\ \cline{2-5} 
                                      & 20    & 50    & 100   & 200   \\ \hline
DFND \cite{chen2021learning}          & 55.25 & 60.25 & 72.87 & 74.78 \\ \hline
$\mathcal{L}_{K\!D}$                  & 60.85 & 64.90 & 72.52 & 74.54 \\
$\;\mathcal{L}_{K\!D}+\mathcal{L}_{AT}$ & 62.50 & 65.66 & 74.48 & 75.95 \\
$\mathcal{L}_{K\!D}+\mathcal{L}_{D}$  & 61.03 & 66.49 & 74.92 & 76.23 \\
$\;\textbf{Full (ours)}$                  & 65.18 & 66.17 & 75.91 & 76.44 \\ \hline
\end{tabular}
}
\vspace{-0.5cm}
\end{table}

\subsection{Diagnostic Experiment}
We first verify the effectiveness of the adaptive threshold $\gamma$. 
We select 300K training samples and conduct 200 epochs uniformly with an average of three training rounds for each combination. 
Simultaneously, the class-dropping rate $\alpha$ is set to 0. 
The experimental results are shown in Table \ref{tab1} (1-5).
The closer $\gamma$ is to 0, the closer it is to the selection method of DFND; the closer to 1, the closer to random selection.
It can be seen from the results that when $\gamma=0.1 $, the sample confidence from the teacher model and the amount of information from the sample can achieve the best combination, which can better help the student model to learn.

We then verify the effect of the class-dropping rate $\alpha$.
Other experimental settings are the same as above, shown in Table \ref{tab1} (6-10).
When $\alpha$ is set to an appropriate value, the accuracy is improved compared with the original method ($\alpha=0$).
For datasets with fewer total classes, a too high class-dropping rate may lose too much information. 
For datasets with larger total classes, a too low class-dropping rate may not be able to suppress label noise effectively. 
So this is why there are differences between the two datasets. In the next experiment, these will also be the default setting for $\gamma$ and $\alpha$.

Finally, we make ablation experiments to verify the effectiveness of each loss. We set different loss combinations in our method, and the DFND \cite{chen2021learning} method. As seen from Table \ref{tab2}, our method has higher accuracy at the 20th epoch than the DFND at the 50th epoch and higher accuracy at the 100th epoch than the DFND at the 200th epoch.
Our method converges faster than the previous state-of-the-art method based on data selection. Through learning structured knowledge, a better student model can be obtained.

\begin{table}[h]
\vspace{-0.3cm}
\centering
\caption{Classification result on the CIFAR dataset. * denotes the results we reproduced using source code and a unified teacher model.}
\label{table3}
\vspace{0.1cm}
\setlength{\tabcolsep}{1mm}
\scalebox{0.8}{
\begin{tabular}{@{}lccc@{}}
\toprule[1pt]
\multirow{2}{*}{Algorithm}                            & \multirow{2}{*}{Extra Costs} &       \multicolumn{2}{c}{Accuracy (\%)} \\ \cmidrule(l){3-4} 
                                              &                           & CIFAR-10                        & CIFAR-100                       \\ \midrule
Teacher                                       & -                         & 95.35                           & 78.60                           \\
Student                                       & -                         & 94.63                           & 77.62                           \\ \midrule
DAFL \cite{chen2019}         & \checkmark & 92.22                           & 74.47                           \\
DFAD \cite{fang2019}         & \checkmark & 93.30                           & 67.70                           \\
DeepInversion \cite{yin2020dreaming} & \checkmark               & 93.26           & -               \\
CMI* \cite{fang2021contrastive}      & \checkmark               & 94.64           & 77.17           \\
ZSKT \cite{micaelli2019zero} & \checkmark & 93.32                           & 67.74                           \\
DFQ* \cite{choi2020data}     & \checkmark & 94.51                           & 77.13                           \\
Fast \cite{fang2021up}       & \checkmark & 94.05                           & 74.34                          \\
DFND \cite{chen2021learning} & $\times$                  & 94.02                           & 76.35                           \\ \midrule
EEIKD-Tiny                                    & $\times$                  & 94.37                           & 76.16                           \\
EEIKD-Large                                   & $\times$                  & \textbf{94.94} & \textbf{77.67} \\ \bottomrule[1pt]
\end{tabular}
}
\vspace{-0.5cm}
\end{table}

\subsection{Comparison to State-of-the-arts}
Table~\ref{table3} shows the results of our proposed EEIKD compared to the state-of-the-art data-free knowledge distillation methods. 
The baselines contain methods that have to require additional computational resources to train additional generators \cite{chen2019,fang2019,yin2020dreaming,fang2021contrastive,micaelli2019zero,choi2020data,fang2021up} and use an unlabeled substitute dataset like DFND \cite{chen2021learning}, which can effectively avoid unnecessary costs. 
From the results, we can see the superiority of our method.
Our unlabeled data selection method can overstep complex generative methods with improved sample utilization efficiency and implicit structured knowledge. 
Although the accuracy of the student is very close to that of the teacher, our method still improves by 0.92\% and 1.32\% compared with the best unlabeled data selection method.

\section{CONCLUSIONS}
\label{sec:typestyle}
In this paper, we first propose an adaptive threshold module to select more effective samples.
Then, a class-dropping mechanism is proposed to suppress label noise between data domains. 
Finally, by adding features distillation between the teacher and student and learning implicit structured relation, the student can converge quickly and learn better, even matching the performance of the trained student network with the original data. 
The accuracy of CIFAR-10 and CIFAR-100 datasets is 94.94$\%$ and 77.67$\%$, respectively, which is better than other data-free knowledge distillation methods.

\footnotesize 
{
\bibliographystyle{IEEEbib}
\bibliography{refs}

\begin{thebibliography}{10}

\bibitem{devlin2018bert}
Jacob Devlin, Ming-Wei Chang, Kenton Lee, and Kristina Toutanova,
\newblock ``Bert: Pre-training of deep bidirectional transformers for language
  understanding,''
\newblock {\em arXiv preprint arXiv:1810.04805}, 2018.

\bibitem{karras2019style}
Tero Karras, Samuli Laine, and Timo Aila,
\newblock ``A style-based generator architecture for generative adversarial
  networks,''
\newblock in {\em Proceedings of the IEEE/CVF conference on computer vision and
  pattern recognition}, 2019, pp. 4401--4410.

\bibitem{yang2023target}
Dingkang Yang, Yang Liu, Can Huang, Mingcheng Li, Xiao Zhao, Yuzheng Wang, Kun
  Yang, Yan Wang, Peng Zhai, and Lihua Zhang,
\newblock ``Target and source modality co-reinforcement for emotion
  understanding from asynchronous multimodal sequences,''
\newblock {\em Knowledge-Based Systems}, p. 110370, 2023.

\bibitem{liu2022learninga}
Yang Liu, Jing Liu, Mengyang Zhao, Dingkang Yang, Xiaoguang Zhu, and Liang
  Song,
\newblock ``Learning appearance-motion normality for video anomaly detection,''
\newblock in {\em 2022 IEEE International Conference on Multimedia and Expo
  (ICME)}. IEEE, 2022, pp. 1--6.

\bibitem{liu2022appearance}
Yang Liu, Jing Liu, Jieyu Lin, Mengyang Zhao, and Liang Song,
\newblock ``Appearance-motion united auto-encoder framework for video anomaly
  detection,''
\newblock {\em IEEE Transactions on Circuits and Systems II: Express Briefs},
  vol. 69, no. 5, pp. 2498--2502, 2022.

\bibitem{Chen_2022_CVPR}
Zhaoyu Chen, Bo~Li, Jianghe Xu, Shuang Wu, Shouhong Ding, and Wenqiang Zhang,
\newblock ``Towards practical certifiable patch defense with vision
  transformer,''
\newblock in {\em Proceedings of the IEEE/CVF Conference on Computer Vision and
  Pattern Recognition}, 2022, pp. 15148--15158.

\bibitem{goodfellow2014generative}
Ian Goodfellow, Jean Pouget-Abadie, Mehdi Mirza, Bing Xu, David Warde-Farley,
  Sherjil Ozair, Aaron Courville, and Yoshua Bengio,
\newblock ``Generative adversarial nets,''
\newblock {\em Advances in Neural Information Processing Systems}, vol. 27,
  2014.

\bibitem{chen2022shape}
Zhaoyu Chen, Bo~Li, Shuang Wu, Jianghe Xu, Shouhong Ding, and Wenqiang Zhang,
\newblock ``Shape matters: deformable patch attack,''
\newblock in {\em Computer Vision--ECCV 2022: 17th European Conference, Tel
  Aviv, Israel, October 23--27, 2022, Proceedings, Part IV}. Springer, 2022,
  pp. 529--548.

\bibitem{liu2022efficient}
Siao Liu, Zhaoyu Chen, Wei Li, Jiwei Zhu, Jiafeng Wang, Wenqiang Zhang, and
  Zhongxue Gan,
\newblock ``Efficient universal shuffle attack for visual object tracking,''
\newblock in {\em ICASSP 2022-2022 IEEE International Conference on Acoustics,
  Speech and Signal Processing (ICASSP)}. IEEE, 2022, pp. 2739--2743.

\bibitem{9428443}
Hao Huang, Yongtao Wang, Zhaoyu Chen, Zhi Tang, Wenqiang Zhang, and Kai-Kuang
  Ma,
\newblock ``Rpattack: Refined patch attack on general object detectors,''
\newblock in {\em 2021 IEEE International Conference on Multimedia and Expo
  (ICME)}, 2021, pp. 1--6.

\bibitem{yang2022disentangled}
Dingkang Yang, Shuai Huang, Haopeng Kuang, Yangtao Du, and Lihua Zhang,
\newblock ``Disentangled representation learning for multimodal emotion
  recognition,''
\newblock in {\em Proceedings of the 30th ACM International Conference on
  Multimedia}, 2022, p. 1642–1651.

\bibitem{yang2022learning}
Dingkang Yang, Haopeng Kuang, Shuai Huang, and Lihua Zhang,
\newblock ``Learning modality-specific and -agnostic representations for
  asynchronous multimodal language sequences,''
\newblock in {\em Proceedings of the 30th ACM International Conference on
  Multimedia}, 2022, p. 1708–1717.

\bibitem{wang2022boosting}
Jiafeng Wang, Zhaoyu Chen, Kaixun Jiang, Dingkang Yang, Lingyi Hong, Yan Wang,
  and Wenqiang Zhang,
\newblock ``Boosting the transferability of adversarial attacks with global
  momentum initialization,''
\newblock {\em arXiv preprint arXiv:2211.11236}, 2022.

\bibitem{9987686}
Zuhao Ge, Lizhe Qi, Yuzheng Wang, and Yunquan Sun,
\newblock ``Zoom-and-reasoning: Joint foreground zoom and visual-semantic
  reasoning detection network for aerial images,''
\newblock {\em IEEE Signal Processing Letters}, vol. 29, pp. 2572--2576, 2022.

\bibitem{huang2022cmua}
Hao Huang, Yongtao Wang, Zhaoyu Chen, Yuze Zhang, Yuheng Li, Zhi Tang, Wei Chu,
  Jingdong Chen, Weisi Lin, and Kai-Kuang Ma,
\newblock ``Cmua-watermark: A cross-model universal adversarial watermark for
  combating deepfakes,''
\newblock in {\em Proceedings of the AAAI Conference on Artificial
  Intelligence}, 2022, vol.~36, pp. 989--997.

\bibitem{wang2023adversarial}
Yuzheng Wang, Zhaoyu Chen, Dingkang Yang, Yang Liu, Siao Liu, Wenqiang Zhang,
  and Lizhe Qi,
\newblock ``Adversarial contrastive distillation with adaptive denoising,''
\newblock {\em arXiv preprint arXiv:2302.08764}, 2023.

\bibitem{lopes2017data}
Raphael~Gontijo Lopes, Stefano Fenu, and Thad Starner,
\newblock ``Data-free knowledge distillation for deep neural networks,''
\newblock {\em arXiv preprint arXiv:1710.07535}, 2017.

\bibitem{chen2019}
Hanting Chen, Yunhe Wang, Chang Xu, Zhaohui Yang, Chuanjian Liu, Boxin Shi,
  Chunjing Xu, Chao Xu, and Qi~Tian,
\newblock ``Data-free learning of student networks,''
\newblock in {\em Proceedings of the IEEE/CVF International Conference on
  Computer Vision}, 2019, pp. 3514--3522.

\bibitem{fang2019}
Gongfan Fang, Jie Song, Chengchao Shen, Xinchao Wang, Da~Chen, and Mingli Song,
\newblock ``Data-free adversarial distillation,''
\newblock {\em arXiv preprint arXiv:1912.11006}, 2019.

\bibitem{micaelli2019zero}
Paul Micaelli and Amos~J Storkey,
\newblock ``Zero-shot knowledge transfer via adversarial belief matching,''
\newblock {\em Advances in Neural Information Processing Systems}, vol. 32,
  2019.

\bibitem{fang2021up}
Gongfan Fang, Kanya Mo, Xinchao Wang, Jie Song, Shitao Bei, Haofei Zhang, and
  Mingli Song,
\newblock ``Up to 100x faster data-free knowledge distillation,''
\newblock {\em arXiv preprint arXiv:2112.06253}, 2021.

\bibitem{yin2020dreaming}
Hongxu Yin, Pavlo Molchanov, Jose~M Alvarez, Zhizhong Li, Arun Mallya, Derek
  Hoiem, Niraj~K Jha, and Jan Kautz,
\newblock ``Dreaming to distill: Data-free knowledge transfer via
  deepinversion,''
\newblock in {\em Proceedings of the IEEE/CVF Conference on Computer Vision and
  Pattern Recognition}, 2020, pp. 8715--8724.

\bibitem{choi2020data}
Yoojin Choi, Jihwan Choi, Mostafa El-Khamy, and Jungwon Lee,
\newblock ``Data-free network quantization with adversarial knowledge
  distillation,''
\newblock in {\em Proceedings of the IEEE/CVF Conference on Computer Vision and
  Pattern Recognition Workshops}, 2020, pp. 710--711.

\bibitem{fang2021contrastive}
Gongfan Fang, Jie Song, Xinchao Wang, Chengchao Shen, Xingen Wang, and Mingli
  Song,
\newblock ``Contrastive model inversion for data-free knowledge distillation,''
\newblock {\em arXiv preprint arXiv:2105.08584}, 2021.

\bibitem{chen2021learning}
Hanting Chen, Tianyu Guo, Chang Xu, Wenshuo Li, Chunjing Xu, Chao Xu, and Yunhe
  Wang,
\newblock ``Learning student networks in the wild,''
\newblock in {\em Proceedings of the IEEE/CVF Conference on Computer Vision and
  Pattern Recognition}, 2021, pp. 6428--6437.

\bibitem{deng2009imagenet}
Jia Deng, Wei Dong, Richard Socher, Li-Jia Li, Kai Li, and Li~Fei-Fei,
\newblock ``Imagenet: A large-scale hierarchical image database,''
\newblock in {\em 2009 IEEE conference on computer vision and pattern
  recognition}. Ieee, 2009, pp. 248--255.

\bibitem{hinton2015distilling}
Geoffrey Hinton, Oriol Vinyals, and Jeff Dean,
\newblock ``Distilling the knowledge in a neural network,''
\newblock {\em arXiv preprint arXiv:1503.02531}, vol. 2, 2015.

\bibitem{yim2017gift}
Junho Yim, Donggyu Joo, Jihoon Bae, and Junmo Kim,
\newblock ``A gift from knowledge distillation: Fast optimization, network
  minimization and transfer learning,''
\newblock in {\em Proceedings of the IEEE conference on computer vision and
  pattern recognition}, 2017, pp. 4133--4141.

\bibitem{ioffe2015batch}
Sergey Ioffe and Christian Szegedy,
\newblock ``Batch normalization: Accelerating deep network training by reducing
  internal covariate shift,''
\newblock in {\em International conference on machine learning}. PMLR, 2015,
  pp. 448--456.

\bibitem{krizhevsky2009learning}
Alex Krizhevsky, Geoffrey Hinton, et~al.,
\newblock ``Learning multiple layers of features from tiny images,''
\newblock 2009.

\bibitem{paszke2019pytorch}
Adam Paszke, Sam Gross, Francisco Massa, Adam Lerer, James Bradbury, Gregory
  Chanan, Trevor Killeen, Zeming Lin, Natalia Gimelshein, Luca Antiga, et~al.,
\newblock ``Pytorch: An imperative style, high-performance deep learning
  library,''
\newblock {\em Advances in neural information processing systems}, vol. 32,
  2019.

\bibitem{he2016deep}
Kaiming He, Xiangyu Zhang, Shaoqing Ren, and Jian Sun,
\newblock ``Deep residual learning for image recognition,''
\newblock in {\em Proceedings of the IEEE conference on computer vision and
  pattern recognition}, 2016, pp. 770--778.

\end{thebibliography}
}
\end{document}